\documentclass{article}
\usepackage{spconf,amsmath,graphicx}

\newcommand\blfootnote[1]{%
  \begingroup
  \renewcommand\thefootnote{}\footnote{#1}%
  \addtocounter{footnote}{-1}%
  \endgroup
}
\newcommand{\methodname}[1]{GROUNDQA}

\usepackage{amsmath,amsfonts,bm}

\def\eqref#1{equation~\ref{#1}}

\def\1{\bm{1}}

\def\ve{{\bm{e}}}

\def\vh{{\bm{h}}}

\DeclareMathAlphabet{\mathsfit}{\encodingdefault}{\sfdefault}{m}{sl}
\SetMathAlphabet{\mathsfit}{bold}{\encodingdefault}{\sfdefault}{bx}{n}

\usepackage{hyperref}
\usepackage{url}
\usepackage{booktabs}
\usepackage{graphicx}

\title{Structured Knowledge Grounding for Question Answering}
\name{Yujie Lu$^{1*}$, Siqi Ouyang$^{1*}$, Kairui Zhou$^2$}
\address{University of California, Santa Barbara$^1$\\Kent School, Connecticut, United States$^2$}
\begin{document}
%
\maketitle

\begin{abstract}
Can language models (LM) ground question-answering (QA) tasks in the knowledge base via inherent relational reasoning ability? While previous models that use only LMs have seen some success on many QA tasks, more recent methods include knowledge graphs (KG) to complement LMs with their more logic-driven implicit knowledge. However, effectively extracting information from structured data, like KGs, empowers LMs to remain an open question, and current models rely on graph techniques to extract knowledge. In this paper, we propose to solely leverage the LMs to combine the language and knowledge for knowledge-based question-answering with flexibility, breadth of coverage, and structured reasoning. Specifically, we devise a knowledge construction method that retrieves the relevant context with a dynamic hop, which expresses more comprehensiveness than traditional GNN-based techniques. And we devise a deep fusion mechanism to further bridge the information exchanging bottleneck between the language and the knowledge. Extensive experiments show that our model consistently demonstrates its state-of-the-art performance over QA benchmarks, including CommensenseQA, showcasing the possibility to leverage LMs solely to robustly ground QA into the knowledge base.\blfootnote{*Equal Contribution.}
\end{abstract}

\section{Introduction}
Question Answering (QA) task: given a question context, a QA model needs to produce an answer for it. This requires the QA model to gather relevant knowledge and reason over it. 
Explicit knowledge source: knowledge graphs (e.g., ConceptNet)

Recent works aim to reason based on both language model and Knoweldge graphs. Language model have been pre-trained on raw texts, storing knowledge explicitly; knowledge graphs are networks connecting entities in a logical way, storing knowledge implicitly. The two systems each have their own role: KGs contains easy-to-follow logic patterns, but lacks coverage and are noisy, thanks to the excessively simplification of logic; LMs vice versa. However, the models differ in how they integrate the two different system together. Starting with QA-GNN, the model runs GNNs over KGs to retrieve a subgraph using entity matching and path finding, then jointly reason over both modalities by either adding question context to the subgraph as an additional node or introducing additional interaction tokens and layers into the model architecture. The drawbacks of using GNN is evident as the retrieved graph is only limited in terms of knowledge as compromised by training performance.

It's difficult to combine these two modes. First, identifying relevant knowledge from a huge KG in the context of a question is difficult. Second, the optimum technique to combine context with the retrieved graph reasoning is unclear. QA-GNN~\cite{Yasunaga2021QAGNNRW} and GreaseLM~\cite{greaseLM}, the most recent work (as far as we know), retrieve a subgraph using entity matching and path finding, then jointly reason over both modalities by either adding question context to the subgraph as an additional node or introducing additional interaction tokens and layers into the model architecture.

Our main contributions are expected as follows:
\begin{itemize}
    \item We propose a new way to fuse these two knowledge bases with only LM involved.
    \item We convert KG data into a natural language format, and then run a LM over it to gain embedding of KG knowledge.
    \item We achieve comparable results to existing SOTA over public QA benchmark CommonsenseQA and OpenbookQA.
\end{itemize}

\section{Methodology}
\subsection{Problem Definition}
For the general natural language question answering task, given question word sequence ${Q = \{q_1,q_2,...,q_{L_Q}\}}$, the model aims at predicting the answer $a \in C$.
${L_Q}$ is the length of the question, and ${q_t}$ denotes a word token that is associated with a ${K}$ dimensional distinct word embedding ${w_t}$.
In the task of knowledge-based context-aware answer selection, the knowledge graph ${G}$ would be added to input to generate answer embedding.

Basically, our method is two step: 1) Use the input sequence to retrieve a subgraph of the knowledge graph G from the knowledge base. 2) Based on the interaction of language and graph, we predict answer a’.

\subsection{Main Approach}
Our initial idea is similar to the QA-GNN one, which also tries to contextualize the QA task into the knowledge graph.
In general, they are trying to use the entity set of the question to retrieve a subgraph $G_s$ from the large Knowledge Graph $G_l$.
And then they convert the QA context to the node and incorporate it into the retrieved subgraph $G_s$, and get the joint graph $G$.
The relevance score between QA context node and Knowledge node are used to represent their dynamic relationship.
Then the GNN module is applied for joint graph representation extraction.
The prediction is based on both the original language model representation and the attention-based context node representation.
After some investigation, we found some limitations of the first idea, such as the knowledge graph is utilized indirectly, and the attention between the QA context and the knowledge requires additional graph construction.
Thus we propose to implement zero-shot QA via a graph-contextualized graph transformer.
Specifically, the language input is attended to the retrieved knowledge subgraph via a GNN-nested transformer architecture.

\subsection{Basic Information Exchange}
We hypothesize that a single interaction token for each modality is insufficient to effectively communicate between text and graph since text sequence is of length $10$-$100$ tokens and graph usually contains hundreds of nodes.
Following GreaseLM, we implement langauge representation and graph representation as follows.
\subsubsection{Language Representation.} ${N + \ell - 1}$ represent the input token embeddings. They are fed into additional transformer LM encoder blocks at each $\ell$-th layer. This process is iterated to encode the textual context based on the LM's pretrained representations:
\begin{align}
    N + \ell =& \text{ LM-Layer}(N + \ell - 1) \label{eq:lm2}\\
    &\text{ for } \ell = 1, \dots, M \nonumber
\end{align}
where $\tilde{\vh}$ corresponds to embeddings of the language modality before the fusion with the graph modality.

\subsubsection{Graph Representation.} The representation of local knowledge subgraph linked from the QA example are encoded by the GNN layers. To represent the graph, following GreaseLM, we first compute initial node embeddings $\{\ve_1^{(0)}, \dots, \ve_J^{(0)}\}$ for the retrieved entities using pretrained KG embeddings for these nodes.
We convert knowledge triples in the KG into sentences by using pre-defined templates for each relation, feeding the sentences into a pretrained BERT-Large LM to get their embeddings, and then perform mean pooling over the tokens of the entity’s occurrences across all the sentences. 
The initial embedding of the interaction node $\ve_{int}^{0}$ is initialized randomly.

In each layer of the GNN, the current representation of the node embeddings $\ell - 1$ is fed into the layer to perform a information propagation between nodes in the graph. In this way, we obtain the node embeddings for each entity before the fusion stage as:
\begin{align}
    \ell =& \text{ GNN}(\ell - 1) \label{eq:gnn} \\
    &\text{ for } \ell = 1, \dots, M \nonumber
\end{align}

\subsubsection{Deep Fusion}
Finally, we need to fuse both language and graph representations together to contextulize our QA into commonsense knowledge base.
Text token will attend to all graph nodes and compute scores. Then scores are transformed into a probability distribution by SoftMax function. Finally the text token receive a weighted sum of graph features.
The details of our cross attention between the language and the graph are summarized in Figure~\ref{fig:deepfusion} and Figure~\ref{fig:attention}.

\begin{table*}[t]
\centering
\resizebox{\textwidth}{!}{%
\begin{tabular}{l c c c c}
\toprule
Model & Naive Fusion & GreaseLM & Cross Attention (1 head) & Cross Attention (4 head) \\
\midrule
IHdev-Acc(\%) & 75.43 & 77.17 & 77.31 & 77.48 \\
IHtest-Acc(\%) & 71.64 & 73.24 & 72.68 & 71.96 \\
\bottomrule
\end{tabular}
}
    \caption{Effects of Deep Fusion.} 
    \label{tab:deep_fusion}
\end{table*}
\begin{table*}[t]
\centering
\resizebox{\textwidth}{!}{%
\begin{tabular}{l c c c c c}
\toprule
Model & No Fusion & GreaseLM & GreaseLM(1-hop) & Cross-Attention & Cross-Attention(1-hop) \\
\midrule
IHdev-Acc(\%) & 75.43 & 77.17 & 77.4 & 77.31 & 78.30 \\
IHtest-Acc(\%) & 71.64 & 73.24 & 72.9 & 72.68 & 72.96 \\
\bottomrule
\end{tabular}
}
    \caption{Effects of the Knowledge Graph Hop.} 
    \label{tab:graph_hop}
\end{table*}
Specifically, let $h^{l-1}_{1:n}$ and $e^{l-1}_{1:m}$ be text features and graph features respectively at layer $l-1$. Following GreaseLM~\cite{greaseLM}, we feed $h$ into a language model layer and $e$ into a GNN layer. Then we have $\tilde{h}^{l-1}_{1:n}$ and $\tilde{e}^{l-1}_{1:m}$. Cross attention is applied on both $\tilde{h}\to\tilde{e}$ and $\tilde{e}\to\tilde{h}$ directions to fuse text and graph information together as illustrated in Figure \ref{fig:deepfusion}.

Take $\tilde{h}\to\tilde{e}$ as an example. Each text token $\tilde{h}^{l-1}_i$ will attend to all graph nodes $\tilde{e}^{l-1}_{1:m}$ independently.
The computation is as follows. For any $1\leq j\leq m$, we have 
\begin{align}
    a_{1,j} &= \tilde{{h}^{l-1}_{1}}^{\top} W_{q}^{\top} W_{k} \tilde{e}^{l-1}_j. \label{eq:sim_score}\\
    \tilde{a}_{1,j} &= \frac{e^{a_{1,j}}}{\sum_{k=1}^m e^{a_{1,k}}} \label{eq:softmax}
\end{align}
Equation \ref{eq:sim_score} calculates the similarity scores of all pairs and \ref{eq:softmax} converts the scores into a probability distribution using softmax function. Finally, graph features are fused into text token by a weighted sum
\begin{align}
    \bar{h}^{l-1}_{1} &= \tilde{h}^{l-1}_1 + W_v \sum_{j=1}^m \tilde{a}_{1,j}\tilde{e}^{l-1}_j \\
    h^l_1 &= \text{MLP}\left(\text{LayerNorm}\left(\bar{h}^{l-1}_{1}\right)\right).
\end{align}

\subsection{Knowledge Seeking}
\subsubsection{Graph Linearization Preprocess}
Given the question $Q$, we ground it into the knowledge graph $G$ and retrieve relevant concept nodes.

\subsubsection{Triplet Representation}
We represent each triplet $T=(t, h, r)$ in the subgraph composed of the retrieved nodes using a sentence-transformer.
Specifically, the embeddings are projected from the node text using the sentence-transformer~\cite{reimers-gurevych-2019-sentence} (RoBERTa-large version released by HuggingFace).

\subsubsection{Context Construction}
Matching the question stem and the triplet.
We keep top-$k$ knowledge triplets with the scoring function below:
\begin{equation}
    Score = \lambda COS(ST([t, h, r]), ST(Q)) + (1-\lambda) RelF(r)
\end{equation}

\section{Experimental Setup}
\subsection{Evaluation Details}
\subsubsection{Datasets and Metrics}
We will mainly evaluate our algorithm on two standard datasets, namely CommonsenseQA~\cite{talmor2018commonsenseqa} and OpenbookQA~\cite{Mihaylov2018CanAS}. CommonsenseQA builds on top of the CONCEPTNET~\cite{liu2004conceptnet}, which contains multiple-choice questions that evaluate whether a model is equipped with different type of common sense. There are $12,247$ questions in total with each question including one correct answer and four distractors. The data is split into a Train/Dev/Test sets with an $80$/$10$/$10$ split. On the other hand,  OpenbookQA contains 5957 4-way multiple choice questions. These questions are based on $1,326$ elementary level science facts. Specifically, there are $4957$/$500$/$500$ questions in the Train/Dev/Test splits.

We will use the test set accuracy as the main metric to evaluate our model when evaluating on both the CommonsenseQA and OpenbookQA datasets.
Specifically, we are using the in-house split for the commonsenseQA dataset.

\subsubsection{Baselines}
We consider the baselines without the knowledge graph~\cite{liu2019roberta, clark2019f, liu2020self} and with the knowledge graph~\cite{schlichtkrull2018modeling, wang2019improving, lin2019kagnet, santoro2017simple, feng2020scalable, Yasunaga2021QAGNNRW, greaseLM}.

\subsection{Implementation Details}
\subsubsection{Model}
We follow GreaseLM~\cite{greaseLM} to use RoBERTa-Large~\cite{liu2019roberta}, AristoRoBERTa~\cite{clark2019f}, and SapBERT~\cite{liu2020self} over CommonsenseQA, OpenbookQA, and MedQA-USMLE dataset respectively to demonstrate the generality in terms of language model initializations.

\subsubsection{Training Configuration}
The basic configuration are following the GreaseLM github repository.
Our experiments are conducted on two A100 GPU for around $5$ hours per experiment.

\subsection{Effects of Deep Fusion}

Cross-modal attention does not provide significant better performance than single-token fusion in GreaseLM.
The results are shown in Table~\ref{tab:deep_fusion}.
One possible reason is that the retrieved graph is too noisy and it’s hard for GNN to extract information from it.

\subsection{Effects of the Knowledge Graph Hop}

Originally, the graph is retrieved by searching for paths of maximum length $3$. To simplify it, we experiment the case of length $1$.
The results are shown in Table~\ref{tab:graph_hop}.
We assume it's because either the GNN is not expressive enough, or the dataset is not large enough to learn a good GNN representation

\section{Discussion}
We propose FuseQA that contextualize the question answering language into the commonsense knowledge base and enables deep fusion of the language and graph modality.
Empirically, the deep cross attention achieves comparable results with the single interaction node method, which indicates that the information bottleneck between two modalities.

In the future, we aim to explore these two directions: 1) How to enable joint reasoning over both the language context and the KG entities without limitation of hop? 2) How to better utilize local and global graph and language information exchange?



\bibliographystyle{IEEEbib}
\bibliography{strings,refs}

\end{document}